\documentclass[10pt,twocolumn,letterpaper]{article}

\usepackage{cvpr}
\usepackage{times}
\usepackage{epsfig}
\usepackage{graphicx}
\usepackage{amsmath}
\usepackage{amssymb}

\usepackage{subcaption}

\usepackage{threeparttable}
\usepackage{multirow, boldline}
\usepackage{booktabs}

\usepackage[ruled,vlined,linesnumbered]{algorithm2e}

\usepackage{authblk}

\makeatletter
\renewcommand\AB@affilsepx{\quad\protect\Affilfont}

\makeatother

\usepackage[pagebackref=true,breaklinks=true,letterpaper=true,colorlinks,bookmarks=false]{hyperref}

\newcommand{\beginsupplement}{%
   \setcounter{table}{0}
   \renewcommand{\thetable}{S\arabic{table}}%
   \setcounter{figure}{0}
   \renewcommand{\thefigure}{S\arabic{figure}}%
}

\cvprfinalcopy 

\def\cvprPaperID{2817} 

\begin{document}

\title{Cross-View Tracking for Multi-Human 3D Pose Estimation at over 100 FPS}


\author[1,2]{Long Chen\thanks{This work was done when Long Chen (long@aifi.com)
was an intern at AiFi Inc.}}
\author[1]{Haizhou Ai}
\author[1,2]{Rui Chen}
\author[1,2]{Zijie Zhuang}
\author[2]{Shuang Liu}

\affil[1]{Department of Computer Science and Technology, Tsinghua University}
\affil[2]{AiFi Inc.}

\maketitle

\begin{abstract}
   Estimating 3D poses of multiple humans
   in real-time is a classic but still challenging task 
   in computer vision.
   Its major difficulty lies 
   in the ambiguity in cross-view association 
   of 2D poses and the huge state space 
   when there are multiple people in multiple views.
   In this paper, we present a novel solution 
   for multi-human 3D pose estimation 
   from multiple calibrated camera views.
   It takes 2D poses in different camera coordinates 
   as inputs and aims for the accurate 3D poses 
   in the global coordinate.
   Unlike previous methods that associate 
   2D poses among all pairs of views 
   from scratch at every frame, 
   we exploit the temporal consistency in videos to 
   match the 2D inputs 
   with 3D poses directly in 3-space. 
   More specifically, we propose to retain the 3D pose
   for each person and update them iteratively 
   via the cross-view multi-human tracking.
   This novel formulation improves both accuracy and efficiency,
   as we demonstrated on widely-used public datasets.
   To further verify the scalability of our method,
   we propose a new large-scale multi-human dataset
   with 12 to 28 camera views.
   Without bells and whistles,
   our solution achieves 154 FPS on 12 cameras
   and 34 FPS on 28 cameras,
   indicating its ability to handle 
   large-scale real-world applications.
   The proposed dataset is released at \url{https://github.com/longcw/crossview_3d_pose_tracking}.



\end{abstract}

\section{Introduction}

Multi-human 3D pose estimation from videos
has a wide range of applications,
including action recognition, sports analysis,
and human-computer interaction.
With the rapid development of deep neural network,
most of the recent efforts in this area have been 
devoted to monocular 3D pose estimation
\cite{moon2019camera,moreno20173d} .
However, despite much progress,
the single-camera setting
is still far from being resolved
due to the large variations of human poses 
and partial occlusion in the monocular views.
A natural solution for these problems 
is to recover the 3D poses
from multiple camera views.


Recent multi-view approaches generally employ
the detected 2D body joints from multiple views as inputs
with the advance of 2D human pose estimation
\cite{cao2017realtime, chen2018cascaded, xiao2018simple},
and address the 3D pose estimation
in a two-step formulation \cite{belagiannis20143d, dong2019fast}.
Specifically,
the 2D joints of 
the same person are first matched and associated across views,
the 3D location of each joint is subsequently
determined by a multi-view reconstruction method.
In this formulation,
the challenge comes from three parts:
1) the detected 2D joints are noisy and inaccurate
since the pose estimation is imperfect;
2) the cross-view association is ambiguous
when multiple people interacting with each other
in crowded scenes;
3) the computational complexity explodes 
as the number of people and number of cameras increase.

To tackle the problem of cross-view association,
3D pictorial structure model (3DPS)
is widely used in some previous methods
\cite{belagiannis20143d, burenius20133d},
where the 3D poses are recovered
from 2D joints in
a discretized 3-space.
In this formulation,
the likelihood of a joint belonging to
a spatial bin is given by 
the geometric consistency 
\cite{hartley2003multiple},
along with a pre-defined body structure model.
A severe problem of 3DPS 
is the expensive computational cost due to 
the huge state space
with multiple people in multiple views.
As an improvement,
Dong \etal \cite{dong2019fast} propose
solving the cross-view association problem 
at the body level 
in advance before applying 3DPS.
They associate 2D poses of the same person from different views
as clusters and estimate 3D poses from the clusters
via 3DPS.
Nevertheless,
matching 2D poses between all pairs of views
still makes the computational complexity explode as 
the number of cameras increases.

\begin{figure}[t]
   \begin{center}
      \includegraphics[width=0.95\linewidth]{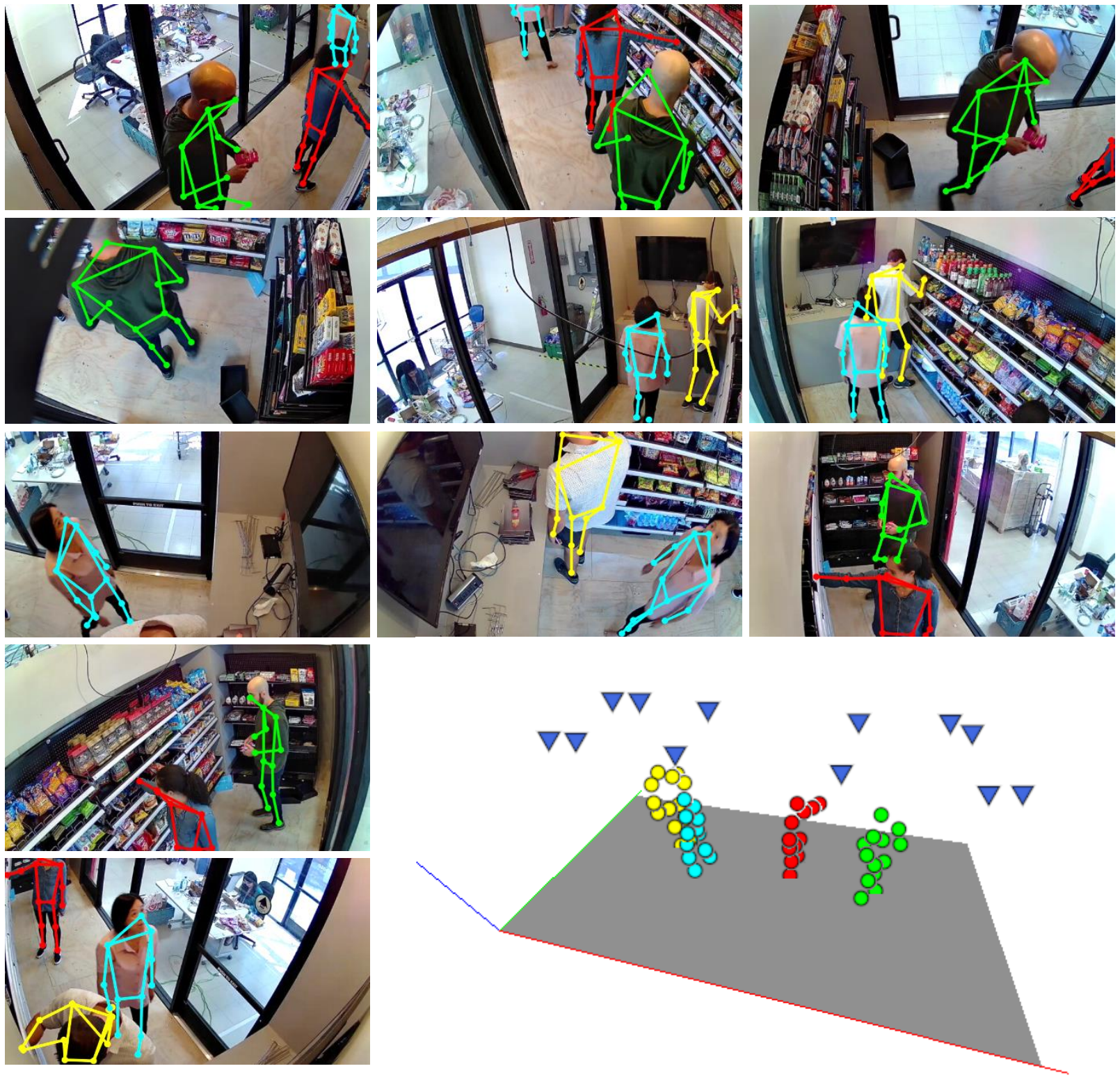}
   \end{center}
   \caption{Multi-human multi-view 3D pose estimation.
   The triangles in the 3D view represent camera locations.}
   \label{fig:multi-view-pose}
\end{figure}

In contrast to previous methods that 
process inputs from multiple cameras simultaneously, 
we propose a new solution 
with an iterative processing strategy.
Specifically, 
we propose exploiting the temporal consistency in videos
to match 2D poses of each view 
with 3D poses directly in 3-space,
where the 3D poses are retained and
updated iteratively
by the cross-view multi-human tracking.
There are two advantages 
in our formulation.
Firstly, for the accuracy,
matching in 3-space is
expected to be robust to partial occlusion
and inaccurate 2D localization,
as the 3D poses consist of multi-view information.
Secondly, for the efficiency,
processing camera views iteratively
makes
the computational complexity
varies only linearly as the number of cameras changes,
enabling the applications on large-scale camera systems.
To verify the effectiveness,
we compare our method with state-of-the-art approaches
on several widely-used
public datasets,
and moreover,
we 
test it on 
a self-collected dataset
with more than 12 cameras,
as shown in Figure~\ref{fig:multi-view-pose}.
With the proposed solution,
we are able to estimate 3D poses
accurately
in 12 cameras at over 100 FPS.

Below, we review related work in
multi-human 3D pose estimation and multi-view tracking,
and then we present the details of our new approach, 
which contains an efficient geometric affinity measurement
for tracking in 3-space,
along with a novel 3D reconstruction algorithm
that designed for iterative processing in videos.
In the experimental section,
we perform the evaluation on 
three public datasets:
Campus \cite{belagiannis20143d},
Shelf \cite{belagiannis20143d},
and CMU Panoptic \cite{joo2015panoptic},
demonstrating both state-of-the-art
accuracy and efficiency of our method.
We also propose a new dataset that
collected from large-scale camera systems,
to verify the scalability of our
method
for real-world applications
as the number of cameras increases.

\section{Related work}
\label{sec:related}

\noindent\textbf{Multi-human 3D pose estimation.}
The problem of 3D human pose estimation
has been studied 
from monocular 
\cite{moreno20173d, andriluka2010monocular,
martinez2017simple, moon2019camera, chengocclusion} 
and multi-view perspectives
\cite{burenius20133d, 
belagiannis2014multiple, dong2019fast, tang2018joint}.

Most of the existing monocular solutions are 
designed for the single-person cases 
\cite{pavllo20193d, martinez2017simple, chengocclusion},
where the estimated poses are
relatively centered around the pelvis joint,
and the absolute locations in the environment are unknown.
Such a relative coordinate setting limits the 
application of these methods in surveillance scenarios.

To estimate multiple 3D poses from a monocular view,
Mehta~\etal \cite{mehta2018single}
use the location-maps
\cite{mehta2017vnect} to infer 
3D joint positions 
at the respective 2D joint pixel locations.
Moon \etal \cite{moon2019camera} 
propose a root localization network
to estimate the camera-centered coordinates 
of the human roots.
Despite lots of recent progress in this area,
the task of monocular 3D pose estimation 
is inherently ambiguous as multiple 3D poses 
can map to the same 2D joints.
The mapping result, unfortunately, often
has a large deviation in practice,
especially when occlusion or motion blur
occurs in images.

On the other hand,
multi-camera systems are becoming
progressively available in the context
of various applications
such as sport analysis and video surveillance.
Given images from multiple camera views,
most previous methods 
\cite{pavlakos2017harvesting, qiu2019cross, 
burenius20133d, belagiannis20143d} 
are generally based on 
the 3D Pictorial Structure model (3DPS) \cite{burenius20133d},
which discretizes the 3-space by an
$N \times N \times N$ grid and 
assigns each joint to one of the $N^3$ bins (hypothesis).
The cross-view association and 
reconstruction are solved
by minimizing the geometric error
\cite{hartley2003multiple} between 
the estimated 3D poses and 2D inputs
among all the hypotheses.
%
%
%
%
Considering all joints of multiple people in
all cameras simultaneously,
these methods are generally computational expensive due to
the huge state space.
Recent work from Dong \etal \cite{dong2019fast}
propose to solve the cross-view association problem
at the body level first.
3DPS is subsequently applied to each cluster of 
the 2D poses of the same person from different views.
The state space is therefore reduced 
as each person
is processed individually.
Nevertheless,
the computational cost of cross-view association
of this method
is still too high to achieve the real-time speed.

\medskip
\noindent\textbf{Multi-view tracking for 3D pose estimation.}
Multi-view tracking for 3D pose estimation
is not a new topic in computer vision. 
However,
it is still nontrivial to combine these two tasks
for fast and robust multi-human 3D pose estimation,
as facing the challenges mentioned above.

Markerless motion capture,
aiming at 3D motion capturing 
for a single person,
has been 
studied for a decade
\cite{taylor2010dynamical, elhayek2015efficient, tome2018rethinking}.
Tracking in these early works
is developed for joint localization
and motion estimation.
As the recent progress in 
deep neural network,
temporal information
is also investigated 
with the recurrent neural network
\cite{rayat2018exploiting, lee2018propagating}
or convolutional neural network
\cite{pavllo20193d}
for single-view 3D pose estimation.
However,
these approaches are generally designed for 
well-aligned
single person cases,
where the critical cross-view association problem
is neglected.

As for the multi-human case,
Belagiannis \etal \cite{belagiannis2014multiple}
propose employing cross-view tracking results
to assist 3D pose estimation 
under the framework of 3DPS.
It introduces the temporal consistency from 
an off-the-shelf
cross-view tracker \cite{berclaz2011multiple}
to reduce the state space of 3DPS.
This approach separates tracking and 
pose estimation into two tasks
and runs at 1 fps,
which is far from being applied 
to the time-critical applications.
There is also a very recent tracking approach
\cite{bridgeman2019multi}
that uses the estimated 3D poses as inputs of the tracker
to improve the tracking quality,
while the pose estimation is rarely benefited
from the tracking results.
Tang \etal \cite{tang2018joint}
propose to jointly perform
multi-view 2D tracking and pose estimation
for 3D scene reconstruction.
The 2D detections are associated using 
a ground plane assumption,
which is efficient but
limits the accuracy.
In contrast,
we couple cross-view tracking and multi-human 3D pose estimation
in a unified framework,
making these two tasks benefit from each other
for both 
accuracy and efficiency.

\section{Method}

In this section, we first give an overview of 
our framework 
with iterative processing,
then we detail the two components of our framework,
that is,
cross-view tracking
in 3-space with geometric affinity measurement
and incremental 3D pose reconstruction in videos.

\subsection{Iterative processing for 3D pose estimation}
\label{sec:overview}

Given an unknown number of people interacting with each other
in the scene covered by multiple calibrated cameras,
our approach takes the detected 2D body joints
as inputs.
We aim at estimating the 3D locations
of a fixed set of body joints for each person in the scene. 
Particularly,
our approach differs from previous methods
in the way they process frames from different cameras.
In contrast to taking all camera views
at a time in a batch mode,
here we assume each camera streams frames independently,
where the frames are collected in chronological order
and fed into the framework one-by-one iteratively.

With iterative processing,
the overall computational cost increases only
linearly as the number of cameras increases,
and the
strict synchronization between cameras 
is no longer required,
making the solution
have the potential to be applied to 
large-scale camera systems.
Such a modification is straightforward,
but not that easy to achieve,
as the cross-view association is generally ambiguous,
especially when only one view
is observed at one time.
Another challenge, in this case, is to 
reconstruct 3D poses from different cameras
when these cameras
are not strictly synchronized.

To solve the problems,
we construct our framework from two components:
1) cross-view tracking for body joint association,
and 2) incremental 3D pose reconstruction
for unsynchronized frames.
Given a frame from a particular camera,
the task of tracking is 
to associate the detected 2D human bodies
with tracked targets.
Here,
we represent the targets in 3-space
using historically estimated 3D poses.
The cross-view association is therefore 
performed between 2D joints and 3D poses
in 3-space,
as detailed in Section~\ref{sec:association}.
Subsequently,
based on the association results,
each 2D human body is assigned 
to a target or labeled as 
unmatched.
The 3D pose of each target
is incrementally updated
when
combining the newly observed and previously retained
2D joints.
Since these joints are from different times,
conventional reconstruction method 
such as triangulation \cite{hartley2003multiple}
is prone to inaccurate 3D locations.
To deal with the unsynchronized frames,
we present our incremental triangulation algorithm 
in Section~\ref{sec:triangulation}.

\subsection{Cross-view tracking with geometric affinity}
\label{sec:association}

In multi-view geometry, 
reconstructing the location of a point in 3-space
requires knowing the 2D locations of the point 
in at least two views. 
Thus in our case,
in order to estimate the 3D poses,
we have to associate the detected 2D joints
across views first.
Similar to \cite{dong2019fast},
we associate the joints at the body level,
but not just across views, also across times.
This forms the cross-view tracking problem,
as discussed in this section.

\medskip
\noindent\textbf{Problem statement.}
We retain historical states of 
persons in the scene as tracked targets,
the problem becomes associating
these targets with
the newly detected human bodies,
while the detections come from
a different camera
in every iteration.
Here,
we begin with some notations and definitions.
We use $\mathbf{x} \in \mathbb{R}^2$ 
to represent 2D point
in camera coordinate,
and $\mathbf{X} \in \mathbb{R}^3$ 
for 3D point in global coordinate.
For a frame from camera $c$ at
time $t$,
a detected human body $D$ is 
denoted as 2D points 
$\mathbf{x}_{t,c}^k $
of a fixed set of human joints 
with indices $k \in \{1,...,K\}$.
Meanwhile,
a target $T$ is represented in 3-space
using points
$\mathbf{X}_{t'}^k \in \mathbb{R}^3$
of the same set of human joints,
where $t'$ 
stands for the last updated time of the joint.
The historical 2D joints are also retained 
in the corresponding targets.

Then, supposing there are  
$M$ detections
$ \{D_{i,t,c} | i = 1, ..., M \}$
in the new frame,
we need to associate
these detections to the last $N$ tracked targets
$ \{T_{i,t'} | i = 1, ..., N \}$,
and afterwards update the 3D locations of targets 
based on the matching results.
Technically, 
this is a weighted bipartite graph matching problem,
where the graph is determined by 
the affinity matrix
$\mathbf{A} \in \mathbb{R}^{N \times M}$
between targets and detections.
Once the graph is determined,
the problem can be solved efficiently with
the Hungarian algorithm
\cite{kuhn1955hungarian}.
Therefore,
our major challenge is to 
measure the affinity
of each pair of targets and detections 
accurately and efficiently.

\medskip
\noindent\textbf{Affinity measurement.}
Given a pair of target and detection
$\langle T_{t'}, D_{t,c} \rangle$,
the affinity is measured from 
both 2D and 3D geometric correspondences:
\begin{equation}
   \label{eq:affinity}
   A(T_{t'}, D_{t,c})= \sum_{k=1}^K 
   A_{2D}(\mathbf{x}_{t'',c}^k,\mathbf{x}_{t,c}^k) 
   + 
   A_{3D}(\mathbf{X}_{t'}^k, \mathbf{x}_{t,c}^k),
\end{equation}
where $\mathbf{x}_{t'',c}^k$
is the last matched joint $k$ of the target
from camera $c$.
For each type of human joints
the correspondence is computed independently,
thus we omit the index $k$
in the following discussion for notation simplicity.

\begin{figure}[t]
   \begin{center}
      \begin{subfigure}[t]{0.22\textwidth}
         \centering
         \includegraphics[width=1\linewidth]{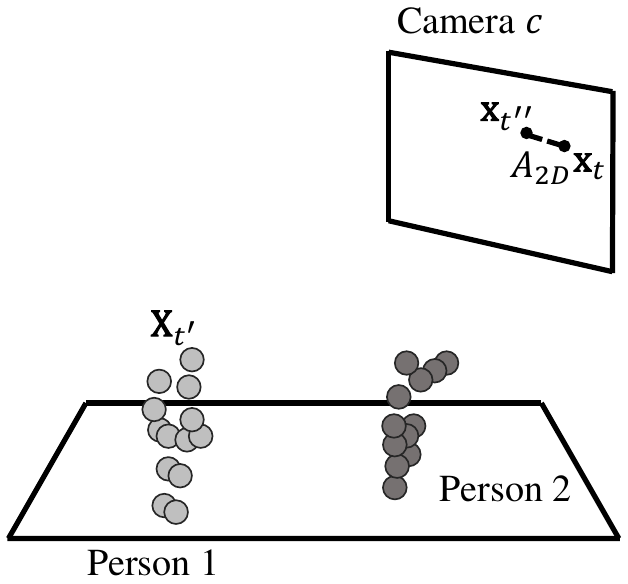}
         \caption{2D correspondence}
         \label{fig:affinity-2d}
      \end{subfigure}%
     ~
   \begin{subfigure}[t]{0.22\textwidth}
      \centering
      \includegraphics[width=1\linewidth]{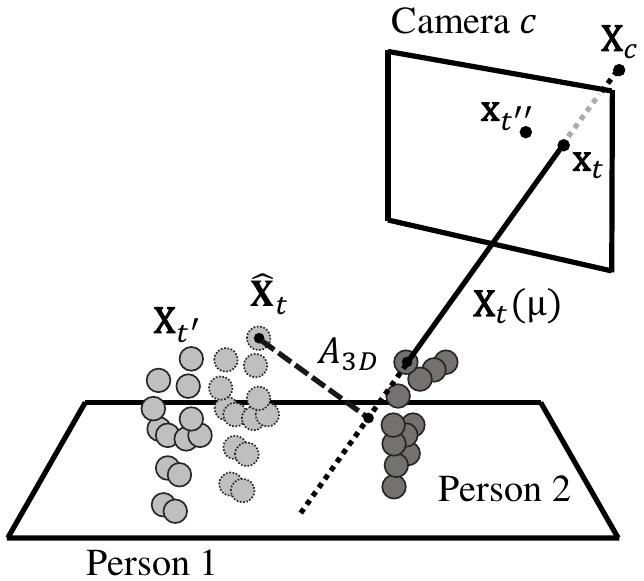}
      \caption{3D correspondence}
      \label{fig:affinity-3d}
  \end{subfigure}%

   \end{center}
   \caption{Geometric affinity measurement.
   (a) 2D correspondence is computed within the same camera.
   (b) 3D correspondence is measured between the predicted location
   and the projected line in 3-space.}
   \label{fig:affinity}
\end{figure}

As shown in Figure~\ref{fig:affinity-2d},
the 2D correspondence
is computed based on the distance of
detected joint $\mathbf{x}_{t,c}$
and 
previously retained joint $\mathbf{x}_{t'',c}$ 
in the camera coordinate:
\begin{equation}
   \label{eq:affinity-2d}
   A_{2D}(\mathbf{x}_{t'',c},\mathbf{x}_{t,c})= w_{2D} (1 - \frac{
      \left \| \mathbf{x}_{t,c} - \mathbf{x}_{t'',c} \right \|
      }{
      \alpha_{2D} (t - t'')
      }) \cdot e^{-\lambda_{a} (t-t'')}.
\end{equation}
There are three hyper-parameters $w_{2D}$, $\alpha_{2D}$,
and $\lambda_a$,
standing for the weight of 2D correspondence,
threshold of 2D velocity, and the penalty rate of time interval,
respectively.
Note that $t > t''$ 
since frames are processed
in chronological order.
$A_{2D} > 0$
indicates these two joints may
come from the same person,
and vice versa.
The magnitude represents the confidence of the indication,
which decreases exponentially 
as the time interval increases.

2D correspondence
is the most basic affinity measurement
that exploited by 
single-view tracking methods.
In order to track people across views,
a 3D correspondence  
is introduced,
as illustrated in Figure~\ref{fig:affinity-3d}.
We suppose that cameras 
are well calibrated
and the
projection matrix of camera $c$
is provided as
$\mathrm{P}_c \in \mathbb{R}^{3 \times 4}$.
We first back-project the detected 2D point 
$\mathbf{x}_{t,c}$
into 3-space as a ray:
\begin{equation}
   \tilde{\mathbf{X}}_{t}(\mu; \mathbf{x}_{t,c}) = 
   \mathrm{P}_c^{+} \tilde{ \mathbf{x} }_{t,c} 
   + \mu \tilde{\mathbf{X}}_c,
\end{equation}
where $\mathrm{P}_c^{+} \in \mathbb{R}^{4 \times 3}$ 
is the pseudo-inverse of $\mathrm{P}_c$
and $\mathbf{X}_c$ is 
the 3D location of the camera center.
The symbol with superscript tilde denotes
the corresponding homogeneous coordinate.
The 3D correspondence is then defined as:
\begin{equation}
   \label{eq:affinity-3d}
   A_{3D}(\mathbf{X}_{t'},\mathbf{x}_{t,c})= w_{3D} (1 - \frac{
      d_l(\hat{\mathbf{X}}_{t} , \mathbf{X}_{t}(\mu))
      }{
      \alpha_{3D}
      }) \cdot e^{-\lambda_a (t-t')},
\end{equation}
where $d_l(\cdot)$ denotes
the point-to-line distance in 3-space
and $\alpha_{3D}$
is the threshold of distance.
Note that in this formulation,
the detected point is compared
with a predicted point $\hat{\mathbf{X}}_{t}$
at the same time $t$.
A linear motion model is introduce
to predict the 3D location at time $t$:
\begin{equation}
   \hat{\mathbf{X}}_{t} = 
   \mathbf{X}_{t'} 
   + \mathbf{V}_{t'} \cdot (t - t'),
\end{equation}
where $t \ge t'$ and $\mathbf{V}_{t'}$ is
3D velocity estimated 
via a linear least-square method.

Here, for the purpose of verifying
the iterative processing strategy,
we only employ the geometric consistency
in the affinity measurement for simplicity.
This baseline formulation already achieves
state-of-the-art performance for both 
human body association and 
3D pose estimation,
as we demonstrated in experiments.
The key contribution comes from 
Equation \ref{eq:affinity-3d},
where we match the detected 2D joints
with targets directly in 3-space.

Compared with 
matching in
pairs of views
in the camera coordinates \cite{dong2019fast},
our formulation has three advantages:
1) matching in 3-space is
robust to partial occlusion
and inaccurate 2D localization,
as the 3D pose 
actually combines the information from multiple views;
2) motion estimation in 3-space 
is more feasible and reliable than that 
in 2D camera coordinates;
3) the computational cost is significantly reduced since 
only one comparison is required
in 3-space for each 
pair of target and detection.
To verify this,
a quantitative comparison is further conducted
in ablation study.


\medskip
\noindent\textbf{Target update and initialization.}
With previous affinity measurement,
this section describes how we update and initialize
targets
in a particular iteration.
Firstly,
we compute the affinity matrix
between targets and detections
using Equation \ref{eq:affinity} 
and solve the association problem
in bipartite graph matching.
Each detection is either assigned to a target
or labeled as unmatched
based on the association results.
In the former case, 
if a detection is assigned to a target, 
the 3D pose of the target will be updated gradually 
with the new detection, 
as the 2D information is observed over time. 
Thus, 3D pose reconstruction in our framework
is an incremental process,
as detailed in Section \ref{sec:triangulation}.

As for the target initialization,
we collect unmatched detections
from different cameras
and associate them
across views using 
epipolar constraint \cite{hartley2003multiple}.
Here for each camera,
only the most recent frame is retained,
thus we assume 
all detections are from 
very similar times 
and can be matched directly.
Particularly,
we solve the association problem
in weighted graph partitioning
\cite{ristani2014tracking, chen2019aggregate},
to comply the cycle-consistency constraint
as there are multiple cameras \cite{dong2019fast}.
Body pose of a new target is
initialized in 3-space from the detections
when at least two views are matched.
The overall procedure of cross-view tracking
is shown in Algorithm \ref{alg:tracking}.

\subsection{Incremental 3D pose reconstruction}
\label{sec:triangulation}
Generally, given 2D poses of the same person
at a time
in different views,
the 3D pose can be 
reconstructed using triangulation.
However,
with the iterative processing,
2D poses in our framework 
may come from different times,
raising the incremental triangulation problem.


Supposing the new frame is from 
camera $c$ at time $t$,
for a target $T_{t'}$ with 
the matched detection $D_{t,c}$
we collect 
2D points
from different cameras
for each type of human joints:
\begin{equation}
   \label{eq:points}
   \mathbf{J}_{t}^k =
   \{ \mathbf{x}_{t,c}^k \}
   \cup 
   \{\mathbf{x}_{t_i,c_i}^k | c_i \neq c \},
\end{equation}
where $\mathbf{x}_{t,c}^k$
is the new point in camera $c$,
and $\mathbf{x}_{t_i,c_i}^k$
denotes the last observed point in camera $c_i$.
For each joint,
the 3D location is estimated independently,
thus we omit the index $k$
in the following discussion for clarity.
Here we aim at estimating the 3D location
$\mathbf{X}_t$ from the
point collection $\mathbf{J}_t$, 
where the points are from different times.

We first briefly introduce the 
linear algebraic triangulation algorithm
and then explain our improvement
that designed for this problem.
For each camera,
the relationship between 2D point
$\mathbf{x}_{t,c}$ 
and 3D point
$\mathbf{X}_t$
can be written as:
\begin{equation}
   \label{eq:projection}
   \tilde{\mathbf{x}}_{t,c} 
   \times 
   ( \mathrm{P}_c \tilde{\mathbf{X}}_t ) = \mathbf{0},
\end{equation}
where $\times$ is the cross product,
$\tilde{\mathbf{x}}_{t,c} \in \mathbb{R}^{3}$
and 
$\tilde{\mathbf{X}}_{t} \in \mathbb{R}^{4}$
are the homogeneous coordinates,
and $\mathrm{P}_c \in \mathbb{R}^{3 \times 4}$
denotes the projection matrix.
Writing Equation \ref{eq:projection} out
on multiple cameras gives
the equation of the form:
\begin{equation}
   \label{eq:linear}
   \mathbf{C} \tilde{\mathbf{X}}_t = \mathbf{0},
\end{equation}
with 
\begin{equation}
   \label{eq:coeff}
   \mathbf{C} = \begin{bmatrix}
      ~~ x_1  \mathbf{p}_{1}^{3\mathrm{T}} - \mathbf{p}_{1}^{1\mathrm{T}} ~~\\ 
      ~~ y_1  \mathbf{p}_{1}^{3\mathrm{T}} - \mathbf{p}_{1}^{2\mathrm{T}} ~~\\ 
      ~~ x_2 \mathbf{p}_{2}^{3\mathrm{T}} - \mathbf{p}_{2}^{1\mathrm{T}} ~~\\ 
      ~~ y_2 \mathbf{p}_{2}^{3\mathrm{T}} - \mathbf{p}_{2}^{2\mathrm{T}} ~~\\ 
      ~~ \ldots ~~
   \end{bmatrix},
\end{equation}
where $(x_c, y_c)$ denotes the 2D point $\mathbf{x}_{t,c}$,
and $\mathbf{p}_{c}^{i\mathrm{T}}$ is 
the $i$-th row of $\mathrm{P}_c$.
If there are at least two views,
Equation~\ref{eq:linear}
is overdetermined
and can be solved via
singular value decomposition (SVD).
The final non-homogeneous coordinate
$\mathbf{X}_t$ 
can be obtained by dividing 
the homogeneous coordinate $\tilde{\mathbf{X}}_t$
by its fourth value:
$\mathbf{X}_t= \tilde{\mathbf{X}}_t / (\tilde{\mathbf{X}}_t)_4$.

The conventional triangulation algorithm
assumes that 2D points of different views 
are from the same time and
independently of each other.
However,
in our case the points are collected 
from different times (Equation \ref{eq:points}).
The time difference between points
varies from 0 to 300 ms in practice,
according to the frame rate and temporary occlusion.

Aiming at estimating the 3D point $\mathbf{X}_t$
for the newest time $t$,
we argue that points from different times
should have different importance 
when solving Equation \ref{eq:linear}.
To this end,
we add weights $\mathbf{w_c}$ to the coefficients
of $\mathbf{C}$ 
corresponding to different cameras:
\begin{equation}
   (\mathbf{w_c} \circ \mathbf{C}) \tilde{\mathbf{X}}_t = \mathbf{0},
\end{equation}
where $\mathbf{w_c} = (w_{1}, w_{2}, w_{3}, w_{4}, ...)$
and $\circ$ denotes Hadamard product.
This is a similar formulation to that in
\cite{iskakov2019learnable},
where $\mathbf{w_c}$ is estimated by a convolution neural network
for the confidences of 2D points.
Differently, our method is designed for
incremental processing on time series:
\begin{equation}
\label{eq:weight}
w_{i} = e^{-\lambda_{t}(t-t_i)} 
/ 
\left \| \mathbf{c}^{i\mathrm{T}} \right \|_2,
\end{equation}
where $\lambda_t$ is the penalty rate,
$t_i \le t$ is the timestamp of the point,
and
$\mathbf{c}^{i\mathrm{T}}$
denotes the $i$-th row of $\mathbf{C}$.
In this case,
the importance of the point 
increases 
as its timestamp closes to the last time,
making the estimated 3D point $\mathbf{X}_t$
closer to the actual joint location at time $t$.
The second term of $L^2$-norm
is written to 
eliminate the bias from
different 2D locations in different views,
as introduced in Equation \ref{eq:coeff}.

\begin{algorithm}[t]
   \caption{Tracking procedure for each iteration}
   \label{alg:tracking}
   \SetAlgoLined
   \DontPrintSemicolon
   \SetNoFillComment
   \footnotesize
   \KwIn{New 2D human poses
   $ \mathbb{D}_{t,c} = \{D_{j,t,c} | j = 1, ..., M \}$ 
   Previous targets 
   $\mathbb{T}_{t'} = \{T_{i,t'} | i = 1, ..., N \}$ at time $t'$
   Previous unmatched detections $\mathbb{D}_\textit{u} = \{ D_{t_i,c_i} \}$}
   \KwOut{New targets with 3D poses
   $ \mathbb{T}_t = \{T_{i,t}\}$ at time $t$}
   
   Initialization: $\mathbb{T}_t \leftarrow \emptyset$; 
   $\mathbf{A} \leftarrow \mathbf{A}_{N \times M} \in \mathbb{R}^{N \times M}$  \;

   \tcc{cross-view association}
   \ForEach{$T_{i, t'} \in \mathbb{T}_{t'}$ }{
      \ForEach{$D_{j,t,c} \in \mathbb{D}_{t,c}$}{
         $\mathbf{A}(i,j) \leftarrow A(T_{i,t'}, D_{j,t,c})$ \;
      }
   }
   $\textit{Indices}_{T},\textit{Indices}_{D} \leftarrow 
   \operatorname{Hungarian Algorithm}(\mathbf{A})$ \;

   \tcc{target update}
   \ForEach{$i, j \in \textit{Indices}_{T},\textit{Indices}_{D}$}{
      $T_{i,t} \leftarrow 
      \operatorname{Incremental 3D Reconstruction}(T_{i,t'}, D_{j,t,c})$ \;
      $\mathbb{T}_t \leftarrow \mathbb{T}_t \cup \{ T_{i,t} \}$ \;
   }

   \tcc{target initialization}
   \ForEach{$j \in \{1,...,M\}$ and $j \notin \textit{Indices}_{D}$}{
      $\mathbb{D}_\textit{u} \leftarrow 
      \mathbb{D}_\textit{u} \cup \{ D_{j,t,c} \} $
   }
   $\mathbf{A}_\textit{u} \leftarrow 
   \operatorname{Epipolar Constraint}(\mathbb{D}_\textit{u})$ \;
   \ForEach{$\mathbb{D}_\textit{cluster} \in 
   \operatorname{Graph Partition}(\mathbf{A}_\textit{u})$}{
      \If{$\operatorname{Length}(\mathbb{D}_\textit{cluster}) \ge 2$}{
         $T_{\textit{new},t} \leftarrow 
         \operatorname{3D Reconstruction}(\mathbb{D}_\textit{cluster})$ \;
         $\mathbb{T}_t \leftarrow \mathbb{T}_t \cup \{ T_{\textit{new},t} \}$ \;
         $\mathbb{D}_\textit{u} \leftarrow 
      \mathbb{D}_\textit{u} - \mathbb{D}_\textit{cluster} $ \;
      }
   }
\end{algorithm}

\section{Experiments}
\label{sec:exp}

We perform the evaluation on 
three widely-used public datasets:
Campus \cite{belagiannis20143d},
Shelf \cite{belagiannis20143d},
and CMU Panoptic \cite{joo2015panoptic},
and
compare our method with 
previous works in terms of 
both accuracy and efficiency.
We also propose a new dataset
with 12 to 28 camera views,
to verify the scalability of our method 
as the numbers of cameras and people increase.

\subsection{Datasets}
We first briefly introduce the public datasets
and evaluation metric for multi-human 3D pose estimation.
Then we present the detail of our proposed dataset
and compare it with existing public datasets.

\textbf{Campus and Shelf.}
The Campus is 
a small-scale dataset that
captured 
by three calibrated cameras.
It consists of three people
interacting with each other 
on an open outdoor square.
The Shelf dataset 
is captured by five cameras 
with a more complex setting,
where four people are interacting 
and disassembling a shelf
in a small indoor area.
The joint annotations of these two 
datasets are provided by Belagiannis \etal
\cite{belagiannis20143d}
for evaluation.
We follow the same evaluation protocol
as in previous works 
\cite{belagiannis20143d, belagiannis20153d, 
ershadi2018multiple, dong2019fast}
and compute the PCP
(percentage of correctly estimated parts)
scores to measure the accuracy
of 3D pose estimation.

\textbf{CMU Panoptic.}
The CMU Panoptic dataset
\cite{joo2015panoptic} is captured
in a closed studio with 480 VGA cameras
and 31 HD cameras.
The hundreds of cameras are 
distributed over the surface of 
a geodesic sphere with about 5 meters of width
and 4 meters of height.
The studio is designed to 
simulate and capture social activities of multiple people
and therefore the space inside the sphere
is built without obstacle.
For the lack of the ground truth 
of 3D poses of multiple people,
only qualitative results 
are presented on this dataset.
In contrast to 
previous works \cite{dong2019fast, iskakov2019learnable} 
that only exploit a few camera views
(about two to five views)
for 3D pose estimation,
we analyze our approach 
with different numbers of cameras
in the ablation study.

\textbf{Our dataset.}
Our dataset, namely Store dataset, is captured inside
two kinds of
simulated stores
with 12 and 28 cameras, respectively.
Different from CMU Panoptic
that uses hundreds of cameras for a small closed area,
we evenly arrange the cameras on the ceiling of
the store 
to simulate the real-world environment.
Each camera works independently without 
strict synchronization,
as we discussed in Section \ref{sec:overview}.
Moreover,
there are lots of shelves inside the second store,
serving as obstacles,
making the scene more complex than
previous datasets.
A detailed comparison is
presented in Table \ref{tab:dataset}.
We use the Store dataset along with 
the CMU Panoptic dataset to verify the 
scalability
of our method on the large-scale camera systems.

\subsection{Comparison with state-of-the-art}
We first present the quantitative comparison
with other state-of-the-art methods 
in Table~\ref{tab:eval}.
Belagiannis \etal
introduced 3DPS for multi-view 
multi-human 3D pose estimation in 
\cite{belagiannis20143d}.
Afterwards, 
they extended 3DPS for videos 
by exploiting the
temporal consistency in \cite{belagiannis2014multiple}.
These early works have a huge state space
with a very expensive computational cost.
Dong \etal \cite{dong2019fast}
propose to cluster
joints at the body level 
to reduce the state space. 
An appearance model \cite{zhong2018camera} 
is also investigated in their work to 
mitigate the ambiguity of the body-level association.
Their approach takes about 25 ms on a dedicated GPU to extract 
appearance features
and 20~ms for the body association, and 60 ms 
for the 3D reconstruction in 3DPS.
Without bells and whistles,
our geometric-only method
outperforms pervious 3DPS-based models
and 
achieves competitive accuracy with
the very recent work \cite{dong2019fast},
while our method is much faster 
with only a single laptop CPU.
Note that, for the fair comparison,
we use the same 2D pose detections for
the experiments as 
that in \cite{dong2019fast},
which are provided by an off-the-shelf 
2D pose estimation method \cite{chen2018cascaded}.

\begin{table}[t]
   \centering
   \resizebox{0.46\textwidth}{!}{%
   \begin{tabular}{r|cccc}
   \hline
   Dataset & Cameras & People & Area & Obstacle \\ \hline
   Campus & 3 & 3 & 43 & None \\
   Shelf & 5 & 4 & 19 & Shelf \\
   CMU Panoptic & 480+31 & 7 & 17 & None \\ \hline
   Store layout1 (ours) & 12 & 4 & 12 & None \\
   Store layout2 (ours) & 28 & 16 & 23 & Shelves \\ \hline
   \end{tabular}%
   }
   \caption{Comparison of datasets.
   The area is computed in square meters using convex hull of camera locations.}
   \label{tab:dataset}
 \end{table}

\begin{table}[t]
   \centering
   \resizebox{0.47\textwidth}{!}{%
   \begin{tabular}{r|cccc|c}
   \hline
   & \multicolumn{4}{c|}{PCP(\%)} &  \\ \cline{2-5} 
   Campus & Actor1 & Actor2 & Actor3 & Average & FPS \\ \hline
   CVPR14 \cite{belagiannis20143d} & 82.0 & 72.4 & 73.7 & 75.8 & - \\
   ECCVW14 \cite{belagiannis2014multiple} & 83.0 & 73.0 & 78.0 & 78.0 & 1 \\
   TPAMI16 \cite{belagiannis20153d} & 93.5 & 75.7 & 85.4 & 84.5 & - \\
   MTA18 \cite{ershadi2018multiple} & 94.2 & 92.9 & 84.6 & 90.6 & - \\
   CVPR19 \cite{dong2019fast} & \textbf{97.6} & 93.3 & 98.0 & 96.3 & 9.5 \\
   \textbf{Ours} & 97.1 & \textbf{94.1} & \textbf{98.6} & \textbf{96.6} & \textbf{617} \\ \hline
   \hline
   Shelf & \multicolumn{1}{l}{Actor1} & \multicolumn{1}{l}{Actor2} & \multicolumn{1}{l}{Actor3} & \multicolumn{1}{l|}{Average} & \multicolumn{1}{l}{FPS} \\ \hline
   CVPR14 \cite{belagiannis20143d} & 66.1 & 65.0 & 83.2 & 71.4 & - \\
   ECCVW14 \cite{belagiannis2014multiple} & 75.0 & 67.0 & 86.0 & 76.0 & 1 \\
   TPAMI16 \cite{belagiannis20153d} & 75.3 & 69.7 & 87.6 & 77.5 & - \\
   MTA18 \cite{ershadi2018multiple} & 93.3 & 75.9 & 94.8 & 88.0 & - \\
   CVPR19 \cite{dong2019fast} & 98.8 & \textbf{94.1} & \textbf{97.8} & \textbf{96.9} & 9.5 \\
   \textbf{Ours} & \textbf{99.6} & 93.2 & 97.5 & 96.8 & \textbf{325} \\ \hline
   \end{tabular}%
   }
   \caption{Quantitative comparison on the Campus and Shelf datasets.
   FPS of other methods is the average speed taken from the papers,
   as per-dataset speed is not provided.}
   \label{tab:eval}
\end{table}

\subsection{Ablation study}
To further verify the effectiveness
of our solution,
ablation study is conducted 
to answer the following questions:
1) Whether matching in 3-space 
has achieved better results
comparing to its 2D counterparts?
2) How much is the contribution of 
the incremental triangulation,
is it really necessary?
3) What is the speed of our method on large-scale camera systems
and how much is the contribution of the iterative processing?
4) How is the quality of the tracking?

\begin{figure}[t]
   \centering
   \includegraphics[width=0.98\linewidth]{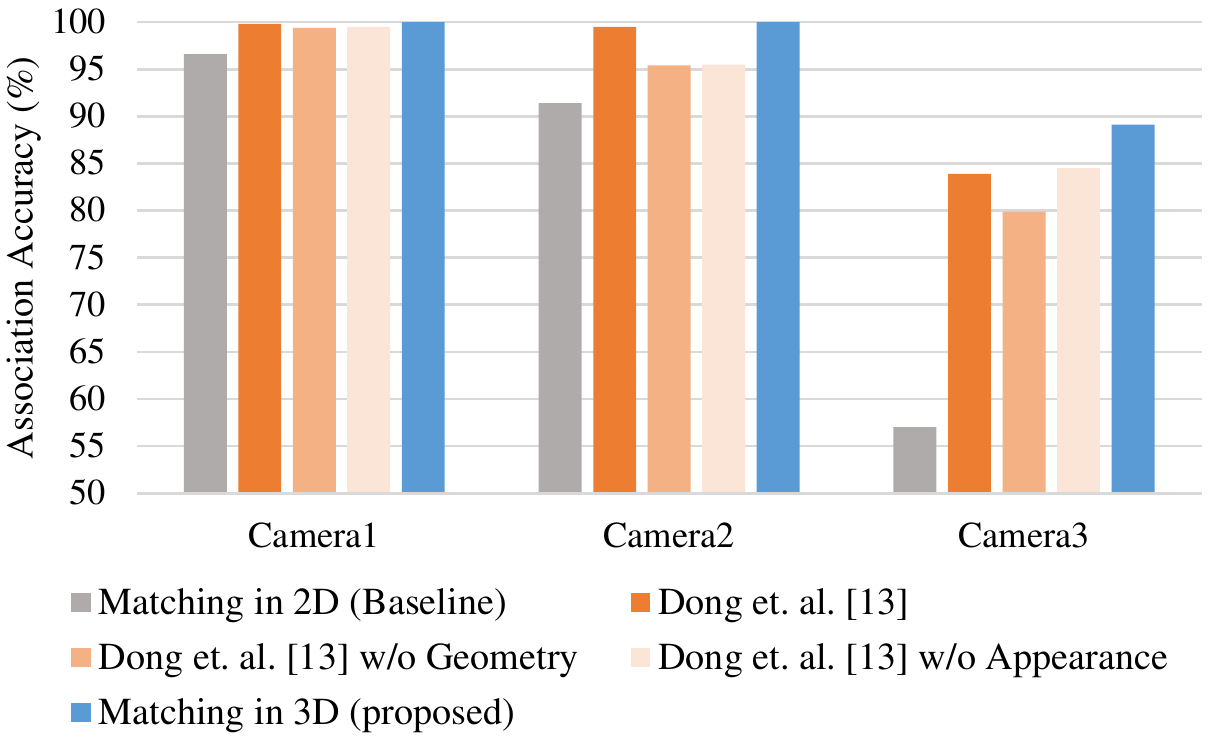}
   \caption{Association accuracy on the Campus dataset.}
   \label{fig:assoication}
\end{figure}

\paragraph{Matching in 2D or 3D?}
As described in Section \ref{sec:association},
we argue that matching in 3-space
leads to more accurate association results,
since it robust to partial occlusion
and inaccurate 2D localization.
To verify that, 
instead of comparing the final 
PCP score,
we measure the association accuracy
directly and compare our method
with four baselines,
as shown in Figure \ref{fig:assoication}.
The association accuracy is 
computed for each camera based on the
degree of agreement between clustered 2D poses and
annotations.
This formulation removes the impact of different 
reconstruction algorithms.
The first baseline is 
matching joints in pairs of views
in the 2D camera coordinates via 
epipolar constraint.
The following three baselines
are taken from the official implementation of
\cite{dong2019fast},
which employs geometric information
and human appearance features
for matching 2D poses between camera views.
As seen in the figure,
all these approaches achieve
good performance in Camera1 and Camera2
of the Campus dataset,
while the gap is revealed 
in the more difficult Camera3,
which is placed closer to the people and 
suffers more from occlusion.
Our method that matching in 3-space 
outperforms the baselines with
32\%, 5.2\%, 9.2\%, 4.6\% association accuracy
in Camera3, respectively.

\paragraph{Different 3D reconstruction methods.}
Cross-view association is the first step of 3D pose estimation,
while 3D reconstruction is also critical.
Here, we retain the association results of our method
and estimate the 3D poses using different 
reconstruction algorithms.
As presented in Table \ref{tab:reconstruction}
four algorithms are considered:
3DPS, conventional triangulation, 
incremental triangulation without normalization,
and our proposed.
We select torso, upper arm,
lower arm for comparison
because these body parts have different motion amplitudes
that can evaluate for different cases.
All the four reconstruction algorithms achieve 
good performance
on the torso as it 
has a small range of motion and 
is easy to detect.
As for the lower arm, which 
can generally move quickly,
our incremental triangulation 
improves about 3\% to 5\% PCP score
compared with conventional triangulation.

\begin{figure}[t]
   \centering
   \includegraphics[width=0.93\linewidth]{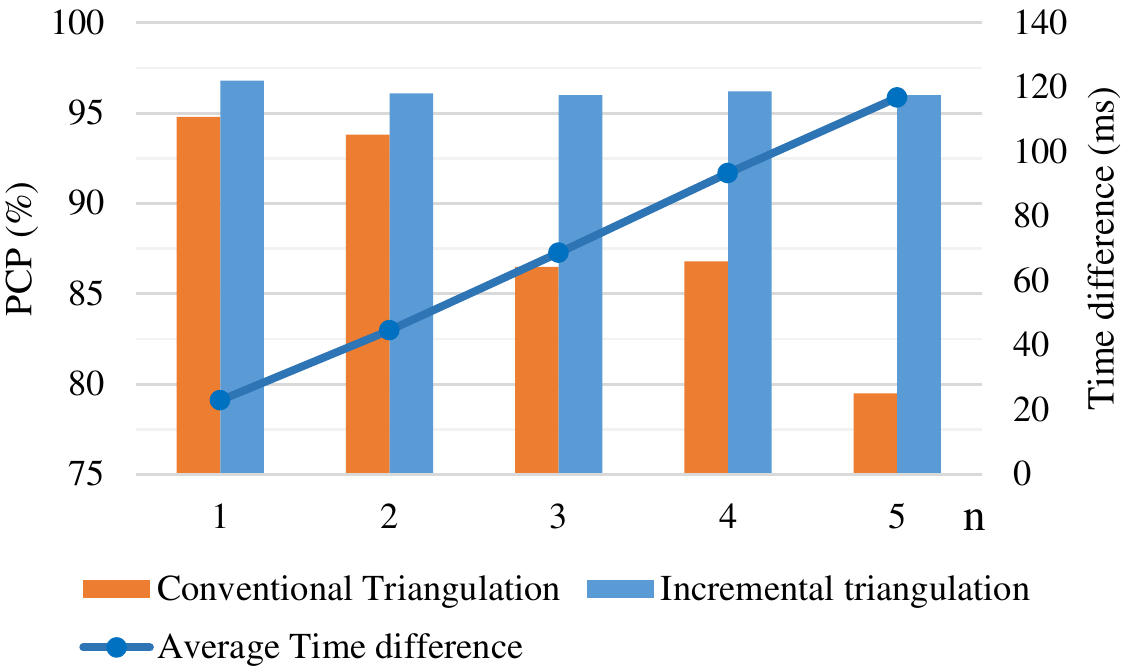}
   \caption{PCP score in terms of input frame rate on the Shelf dataset.
   The original frame rate is 25,
   therefore the actual frame rate of each trail is $25 / n$.}
   \label{fig:time-diff}
\end{figure}

To further verify if the incremental triangulation
has the ability to handle unsynchronized frames,
we analyze the performance drop
when the input frame rate decreases.
The original Shelf dataset was captured 
with 25 FPS.
We construct datasets with different frame rates
by sampling one frame from every $n$ frames 
in each camera.
The comparison between incremental
and conventional triangulation
is shown in Figure \ref{fig:time-diff}.
Average time differences 
within every 2D joint collection $\mathbf{J}_t^k$
are also recorded in the figure.
As the input frame rate decreases
and the time differences increase,
the performance of conventional triangulation 
drops significantly,
while that of ours keeps stable,
indicating the effectiveness of our method
in handling the unsynchronized frames.
Therefore,
we confirm that 
incremental triangulation is 
essential
for the iterative processing. 


\begin{table}[t]
   \centering
   \resizebox{0.47\textwidth}{!}{%
   \begin{tabular}{r|cccc}
   \hline
   Campus & Torso & Upper arm & Lower arm & Whole \\ \hline
   3DPS & 100.0 & \textbf{99.1} & 82.5 & 96.0 \\
   Triangulation & 100.0 & 95.4 & 79.1 & 94.4 \\
   \textbf{Ours, w/o norm} & 100.0 & 95.6 & 81.7 & 95.4 \\
   \textbf{Ours, proposed} & 100.0 & 98.6 & \textbf{84.6} & \textbf{96.6} \\ \hline
   \hline
   Shelf & Torso & Upper arm & Lower arm & Whole \\ \hline
   3DPS &  100.0 & 98.1 & \textbf{88.4} & 96.6 \\
   Triangulation & 100.0 & 97.0 & 84.5 & 94.8 \\
   \textbf{Ours, w/o norm} & 100.0 & \textbf{98.7} & 87.7 & \textbf{96.9} \\
   \textbf{Ours, proposed} & 100.0 & \textbf{98.7} & 87.7 & 96.8 \\ \hline
   \end{tabular}%
   }
   \caption{PCP scores of different 3D reconstruction algorithms
   on the Campus and Shelf datasets.}
   \label{tab:reconstruction}
\end{table}

\begin{figure*}
   \begin{center}
      \includegraphics[width=0.96\linewidth]{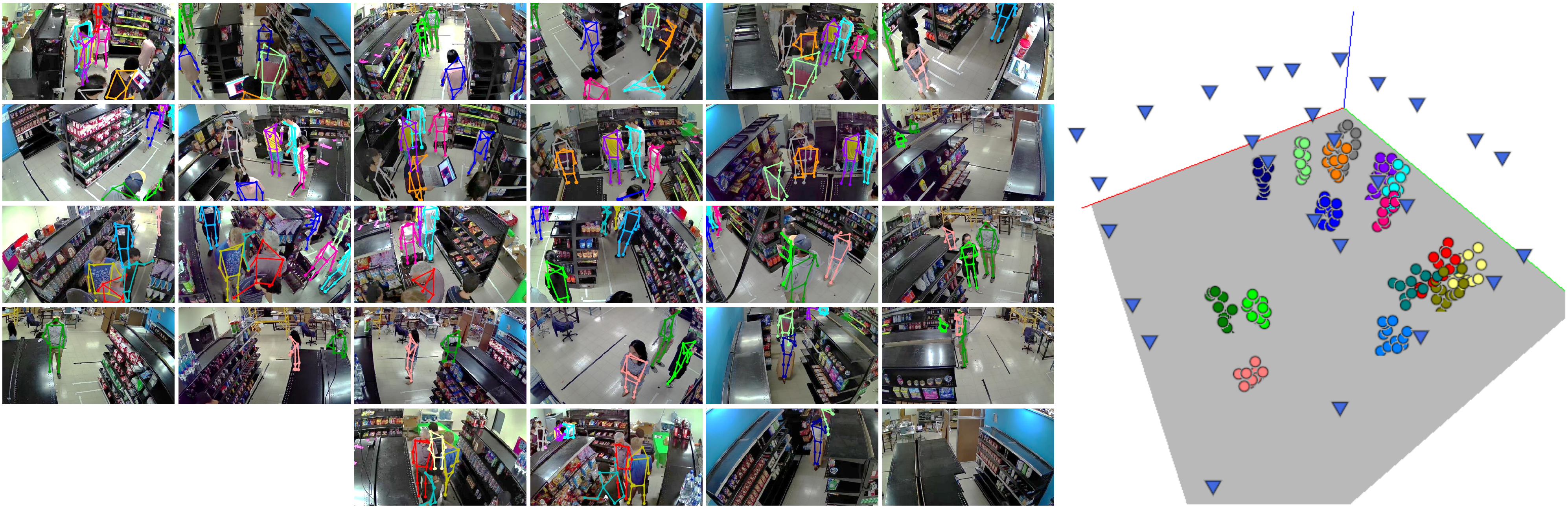}
   \end{center}
   \caption{Qualitative result on the Store dataset (layout 2). 
   There are 28 cameras and 16 people in the scene 
   and different people are represented in different colors.
   The camera locations are illustrated in the 3D view as triangles in blue.}
   \label{fig:demo}
\end{figure*}

\paragraph{Speed on large-scale camera system.}
As already seen in Table \ref{tab:eval},
our method is about 50 times faster
than others
on the small-scale datasets Campus and Shelf.
We further test the proposed method 
on the large-scale Store dataset
as demonstrated in Figure \ref{fig:demo}.
It finally achieves 154 FPS for 12 cameras
with 4 people and 
34 FPS for 28 cameras with 16 people.
Note that 
when counting the running speed,
we follow the common practice that
one frame represents that all cameras are updated once.

\begin{figure}[t]
   \centering
   \includegraphics[width=0.94\linewidth]{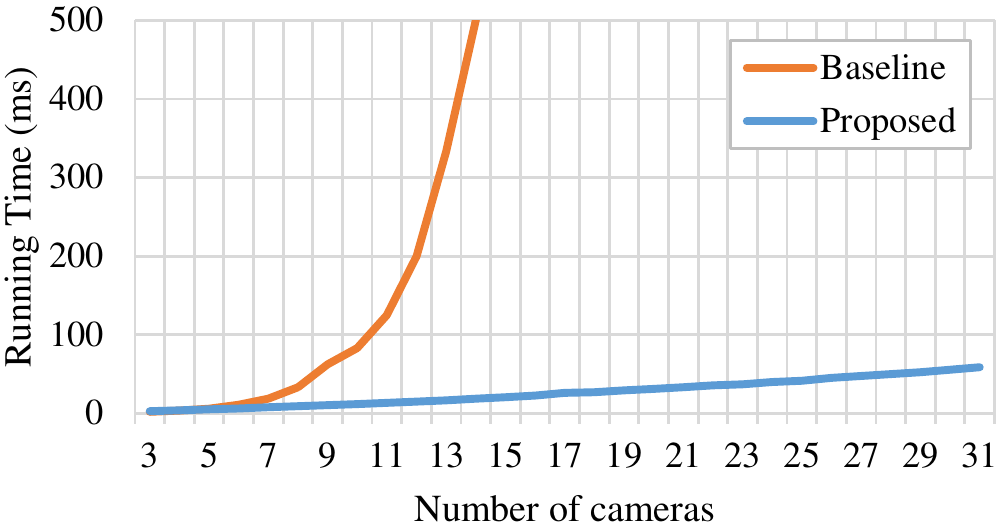}
   \caption{Average running time of one frame with different numbers of cameras
   on the CMU Panoptic dataset.}
   \label{fig:speed}
\end{figure}

Indeed, different implementation and
hardware environment
affect the running speed a lot.
Our algorithm is implemented in C++
without multi-processing
and evaluated on the laptop
with an Intel i7 2.20 GHz CPU.
In order to verify the efficiency
more fairly
and understand the contribution of iterative processing,
we construct a baseline method
that matches joints 
in pairs of views in the camera coordinates
with the same testing environment.
The comparison is conducted on the CMU Panoptic
dataset with its 31 HD cameras,
as the cameras are all placed in a closed small area that
changing the number of cameras does not affect
the number of people observed.
As shown in Figure \ref{fig:speed},
the running time of the baseline method
explodes as the number of cameras increases,
while that of ours varies almost linearly.
The result verifies the 
effectiveness of the iterative processing strategy
and
demonstrates the ability of our method
to work with large-scale camera systems
in real-world applications.

\paragraph{Tracking quality.}
The quality of tracking is measured in each camera view
using the Shelf dataset.
Particularly,
we project the estimated 3D poses onto each camera and
follow the same evaluation protocol as MOTChallenge 
\cite{milan2016mot16}.
We compare our result with 
a simple single-view tracking baseline \cite{Bewley2016_sort}
as shown in Table~\ref{tab:tracking}.
In some easy cases, e.g.\ Camera2 and Camera3,
the baseline single-view tracker achieves similar performance
as cross-view tracking.
But for the difficult cases
such as Camera4 and Camera5, which contain
severe occlusion,
our cross-view tracking outperforms its single-view counterpart 
significantly.
The experimental result verifies that, in our framework,
multi-human tracking can also
be boosted by 
multi-view 3D pose estimation.

\begin{table}[t]
   \centering
   \resizebox{0.45\textwidth}{!}{%
   \begin{tabular}{r|c|ccccc}
   \hline
   Method      & Camera  & MOTA & IDF1 & FP & FN & IDS \\ \hline
   Single-view & Camera1 & 86.7 & 81.7 & 32 & 34 & 2   \\
               & Camera2 & 97.6 & 63.9 & 4  & 4  & 4   \\
               & Camera3 & 97.3 & 98.6 & 7  & 7  & 0   \\
               & Camera4 & 68.8 & 41.8 & 77 & 79 & 3   \\
               & Camera5 & 79.0 & 69.0 & 51 & 51 & 5   \\ 
               \hline
               \hline
   Cross-view  & Camera1 & 98.8 & 99.4 & 3  & 3  & 0   \\
               & Camera2 & 99.2 & 99.6 & 1  & 1  & 2   \\
               & Camera3 & 98.4 & 99.2 & 4  & 4  & 0   \\
               & Camera4 & 97.6 & 98.8 & 6  & 6  & 0   \\
               & Camera5 & 97.6 & 98.8 & 6  & 6  & 0   \\ \hline
   \end{tabular}%
   }
   \caption{Tracking performance on the Shelf dataset.}
   \label{tab:tracking}
\end{table}

\section{Conclusion}
We have presented a novel solution for 
multi-human 3D pose estimation from multiple camera views.
By exploiting the temporal consistency in videos,
we propose to match the 2D inputs with 3D poses
in 3-space directly,
where the 3D poses are retained and
iteratively updated 
by a cross-view tracking.
In experiments,
we have achieved state-of-the-art accuracy 
and efficiency on three public datasets.
The comprehensive ablation study 
demonstrates the effectiveness 
of each component in our framework.
Given
its simple formulation and efficiency, 
our solution can be
extended easily by other techniques 
such as appearance features, 
and applied directly to other high-level tasks.
In addition, 
we propose a new large-scale Store dataset 
to simulate the real-world scenarios, 
which verifies the scalability of our solution 
and may also benefit future researches in this area.

\section{Supplementary Material}
\beginsupplement

\subsection{Detail of Target Initialization}
Here,
we present details of 
our target initialization algorithm,
including the epipolar constraint,
cycle-consistency,
and the formulation we utilized
for graph partitioning.

\begin{figure}[t]
    \centering
    \includegraphics[width=0.71\linewidth]{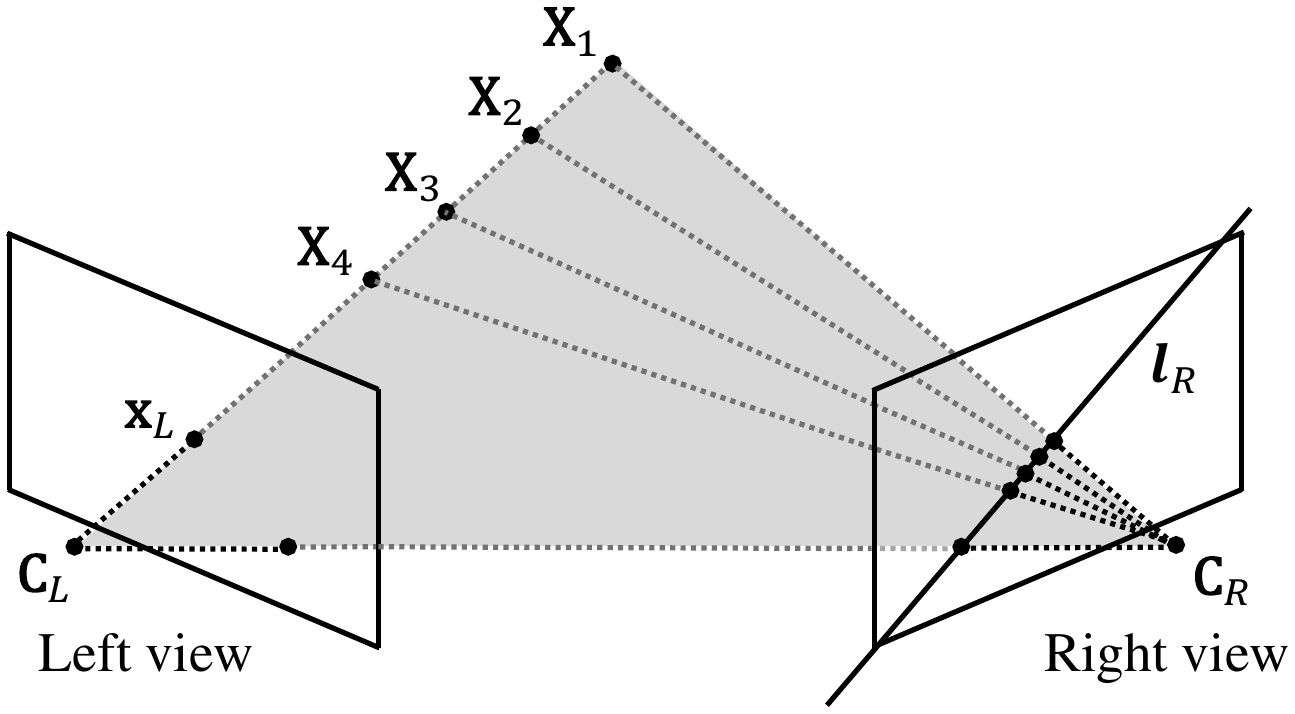}
    \caption{Epipolar constraint: 
    given $\mathbf{x}_L$,
    the projection
    on the right camera plane $\mathbf{x}_R$
    must be on the epipolar line $l_R$.}
    \label{fig:epipolar}
 \end{figure}

When two cameras observing a 3D point 
from two distinct views,
the epipolar constraint \cite{hartley2003multiple} provides 
relations between the two projected 2D points
in camera coordinates,
as illustrated in Figure \ref{fig:epipolar}.
Supposing $\mathbf{x}_L$ is the projected 2D point
in the left view, 
the another projected point $\mathbf{x}_R$ of the right view
should be contained in the epipolar line:
\begin{equation}
    l_R = \mathrm{F}\mathbf{x}_L,
\end{equation}
where $\mathrm{F}$ is the fundamental matrix
that determined by the internal parameters and 
relative poses of the two cameras.
Therefore,
given two points from two views,
we can measure the correspondence between them
based on the point-to-line distance 
in the camera coordinates:
\begin{equation}
    \label{eq:epipolar}
    A_{e}(\mathbf{x}_L, \mathbf{x}_R) = 
1 - \frac{d_{l}(\mathbf{x}_L, l_L) + d_l(\mathbf{x}_R, l_R)}{2 \cdot \alpha_{2D}}.
\end{equation}

Given a set of unmatched detections
$\{D_i \}$ from different cameras,
we compute the affinity matrix
using Equation \ref{eq:epipolar}.
Then the problem is turned to associate these detections across
camera views.
Note that there are multiple cameras,
the association problem can not be formulated as
simple bipartite graph partitioning.
And the matching result should satisfy the cycle-consistent constraint,
i.e.\ $\langle D_i, D_k  \rangle$
must be matched if 
$\langle D_i, D_j  \rangle$
and 
$\langle D_j, D_k  \rangle$
are matched.
To this end,
we formulate the problem as 
general graph partitioning
and solve it via binary integer programming
\cite{chen2019aggregate, ristani2014tracking}:
\begin{equation}
    \label{eq:bip}
    \mathbf{y}^* = \operatorname*{argmax}_Y
  \sum_{\langle D_i, D_j \rangle}
  a_{ij}y_{ij},
\end{equation}
subject to
\begin{equation}
    y_{ij} \in \{0,1\} ,
\end{equation}
\begin{equation}
    \label{eq:transitive}
    y_{ij}+y_{jk} \leq 1 + y_{ik} ,
\end{equation}
where $a_{ij}$ is the affinity between $\langle D_{i}, D_{j} \rangle$
and $Y$ is the set of all possible 
assignments to the binary variables $y_{ij}$.
The cycle-consistency constraint is 
ensured by Equation \ref{eq:transitive}.

\subsection{Baseline Method in the Ablation Study}
To verify the effectiveness of 
our solution,
we construct a method that 
matches joints in pairs of views
using epipolar constraint
as the baseline in ablation study.
The procedure of the baseline method is
detailed in Algorithm \ref{alg:baseline}.
Basically,
for each frame,
it takes 2D poses from all cameras as inputs,
and associate them across views using
epipolar constraint and graph partitioning.
Afterwards, 3D poses are estimated from
the matching results via triangulation.

\begin{algorithm}[b]
    \SetAlgoLined
    \DontPrintSemicolon
    \SetNoFillComment
    \footnotesize
    \KwIn{2D human poses
    $ \mathbb{D} = \{D_{i,c_i} | i = 1, ..., M \}$ }
    \KwOut{3D poses of all people
    $ \mathbb{T} = \{T_{i}\}$}
    
    Initialization: $\mathbb{T} \leftarrow \emptyset$; 
    $\mathbf{A} \leftarrow \mathbf{A}_{M \times M} \in \mathbb{R}^{M \times M}$  \;
 
    \ForEach{$D_{i,c_i} \in \mathbb{D}$ }{
       \ForEach{$D_{j,c_j} \in \mathbb{D}$}{
           \uIf{$c_i \neq c_j$}{
            $\mathbf{A}(i,j) \leftarrow A_{e}(D_{i,c_i}, D_{j,c_j})$ \;
           }
           \Else{
            $\mathbf{A}(i,j) \leftarrow -\inf$ \;
           }
          
       }
    }

    \ForEach{$\mathbb{D}_\textit{cluster} \in 
    \operatorname{Graph Partitioning}(\mathbf{A})$}{
       \If{$\operatorname{Length}(\mathbb{D}_\textit{cluster}) \ge 2$}{
          $\mathbb{T} \leftarrow \mathbb{T} \cup 
          \operatorname{Triangulation}(\mathbb{D}_\textit{cluster})$ \;
       }
    }
    \caption{Baseline for 3D pose estimation.}
    \label{alg:baseline}
 \end{algorithm}

\subsection{Parameter Selection}

In this work, we have six parameters:
$w_{2D}$, $w_{3D}$ are the weights of 
the affinity measurements,
$\alpha_{2D}$ and $\alpha_{3D}$
are the corresponding thresholds,
and $\lambda_a$, $\lambda_t$
are the time penalty rates
for the affinity calculation
and incremental triangulation, respectively.
Here in Table \ref{tab:weights}, 
we first show the experimental results
with different affinity weights
on the Campus dataset.
As seen in the table,
3D correspondence is critical
in our framework
but the performance is robust to 
the combination of weights.
Therefore, 
we fix $w_{2D}=0.4$,
$w_{3D}=0.6$ for all datasets,
and select other parameters for each dataset empirically,
as shown in Table \ref{tab:parameter}.
The basic intuition behind it 
is to adjust $\alpha_{2D}$ according to 
the image resolution and 
change $\lambda_a$, $\lambda_t$ 
based on the input frame rate.
Since different datasets are 
captured at different frame rates, 
e.g. the first three public datasets 
are captured at 25 FPS
while the Store dataset is captured at 10 FPS.


\begin{table}[h]
    \centering
    \resizebox{0.43\textwidth}{!}{%
    \begin{tabular}{cc|cc}
    \hline
    $w_{2D}$ & $w_{3D}$ & Association Accuracy (\%) & PCP (\%) \\ \hline
    1.0 & 0.0 & 45.69 & 62.29 \\
    0.8 & 0.2 & 96.22 & 96.58 \\
    0.6 & 0.4 & 96.30 & 96.61 \\
    \textbf{0.4} & \textbf{0.6} & \textbf{96.38} & \textbf{96.63} \\
    \textbf{0.2} & \textbf{0.8} & \textbf{96.38} & \textbf{96.63} \\
    0.0 & 1.0 & 96.38 & 96.49 \\
    \hline
    \end{tabular}%
    }
    \caption{Association accuracy and PCP score 
    with different weight combinations
    on the Campus dataset.}
    \label{tab:weights}
\end{table}

\begin{table}[h]
    \centering
    \resizebox{0.47\textwidth}{!}{%
    \begin{tabular}{r|cccc}
    \hline
    Dataset & $\alpha_{2D}$ \textit{(pixel / second)} & $\alpha_{3D}$ \textit{(m)} & $\lambda_a$ & $\lambda_t$ \\ \hline
    Campus & 25 & 0.10 & 5 & 10 \\
    Shelf & 60 & 0.15 & 5 & 10 \\
    CMU Panoptic & 60 & 0.15 & 5 & 10 \\
    Store (layout 1) & 70 & 0.25 & 3 & 5 \\
    Store (layout 2) & 70 & 0.25 & 3 & 5 \\ \hline
    \end{tabular}%
    }
    \caption{Parameter selection for each dataset.}
    \label{tab:parameter}
\end{table}



\subsection{Qualitative Results}
\begin{figure}[t]
    \begin{center}
       \includegraphics[width=0.92\linewidth]{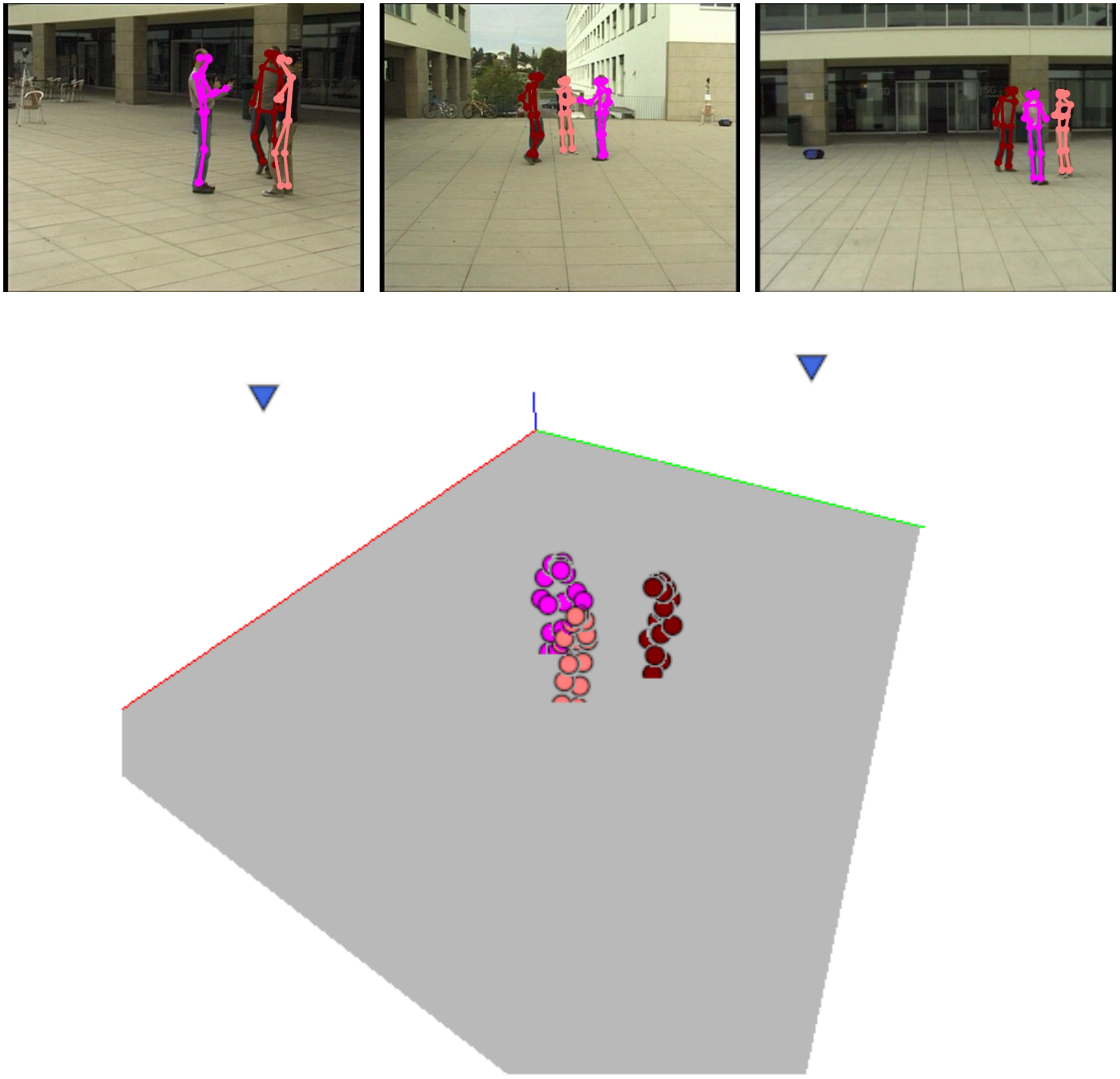}
    \end{center}
    \caption{Qualitative result on the Campus dataset. 
    There are three people with three cameras in an outdoor square.
    Different people are represented in different colors
    based on the tracking result.
    The camera locations are illustrated in the 3D view as triangles in blue.}
    \label{fig:campus}
 \end{figure}

\begin{figure}[t]
    \begin{center}
       \includegraphics[width=0.93\linewidth]{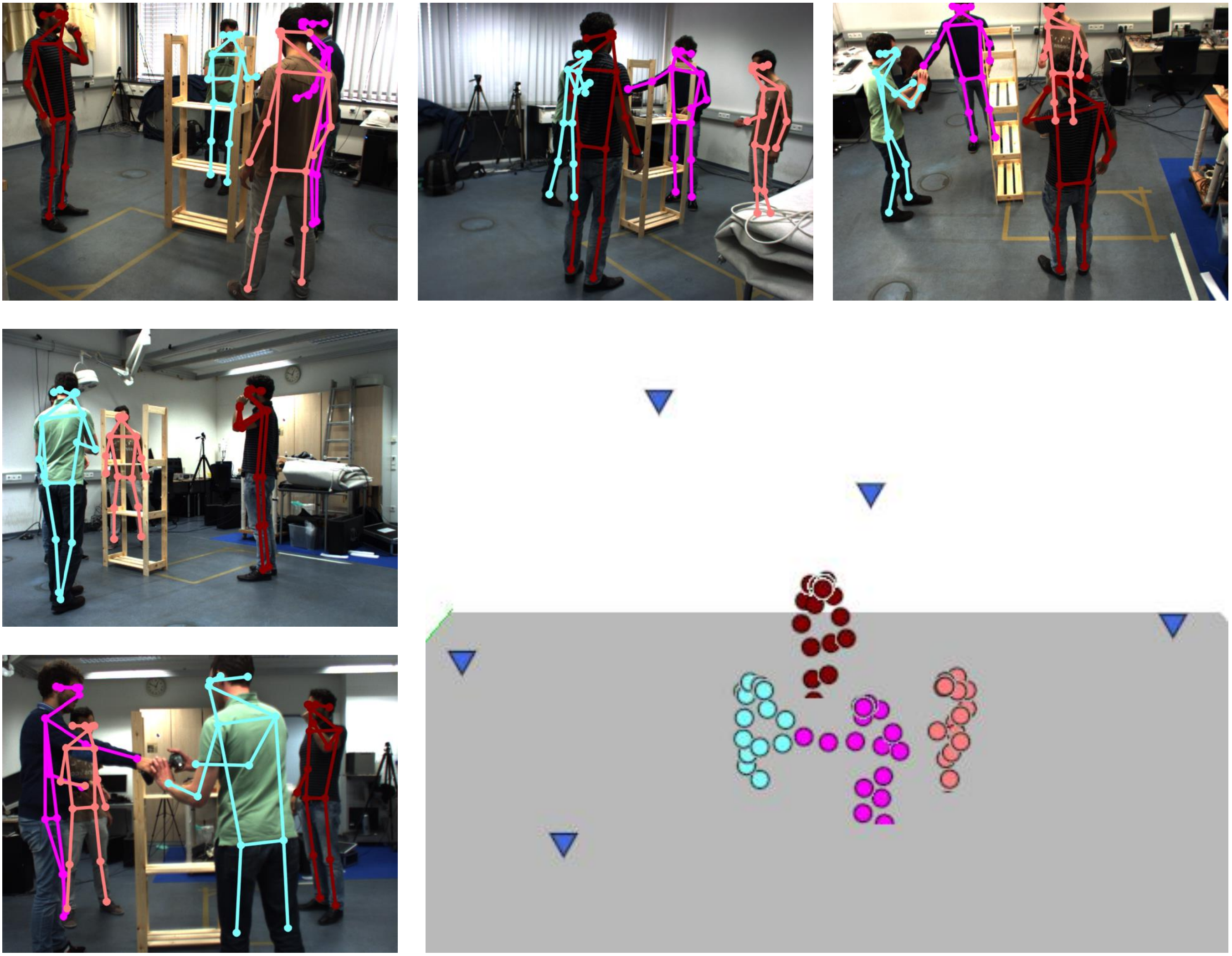}
    \end{center}
    \caption{Qualitative result on the Shelf dataset.
    It consists of
    four people disassembling a shelf under five cameras.
    The camera locations are illustrated in the 3D view as triangles in blue.
    The actions of people can be seen clearly from the estimated 3D poses.
    }
    \label{fig:shelf}
 \end{figure}

\begin{figure*}[t]
    \begin{center}
       \includegraphics[width=0.8\linewidth]{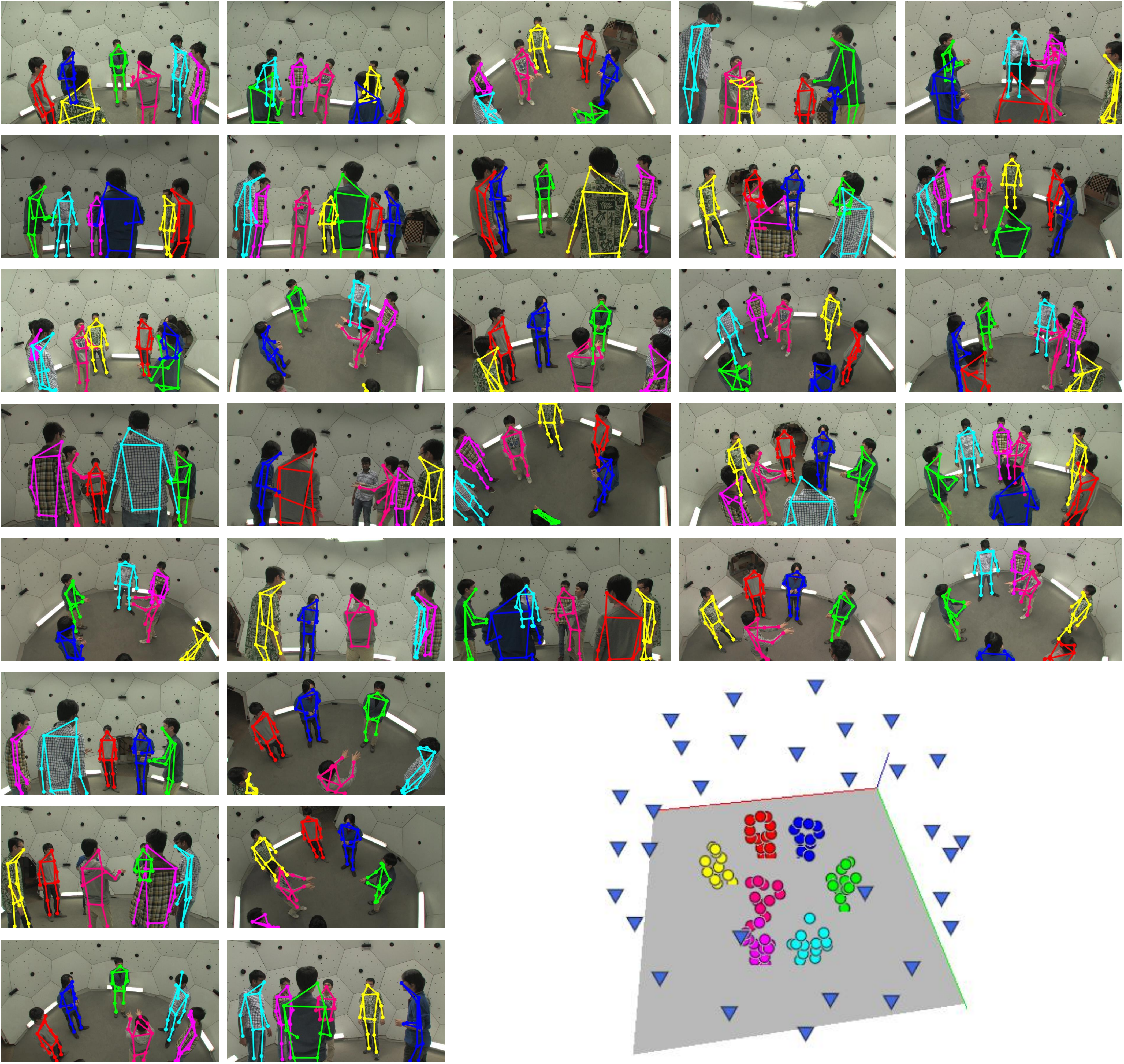}
    \end{center}
    \caption{Qualitative result on the CMU Panoptic dataset. 
    There are 31 cameras and 7 people in the scene.
    The cameras are distributed over the surface of 
    a geodesic sphere.
    As we detailed in ablation study,
    with the proposed iterative processing all the 31 cameras can be
    updated in around 0.058 seconds.}
    \label{fig:cmu}
 \end{figure*}

Here, we present more qualitative results
of our solution
on public datasets
in Figure \ref{fig:campus}, Figure \ref{fig:shelf}, and Figure \ref{fig:cmu}.
A recorded video is also provided at
\url{https://youtu.be/-4wTcGjHZq8}.

{\small
\bibliographystyle{ieee_fullname}
\bibliography{egbib}

\begin{thebibliography}{10}\itemsep=-1pt

\bibitem{andriluka2010monocular}
Mykhaylo Andriluka, Stefan Roth, and Bernt Schiele.
\newblock Monocular 3d pose estimation and tracking by detection.
\newblock In {\em 2010 IEEE Computer Society Conference on Computer Vision and
  Pattern Recognition}, pages 623--630. IEEE, 2010.

\bibitem{belagiannis20143d}
Vasileios Belagiannis, Sikandar Amin, Mykhaylo Andriluka, Bernt Schiele, Nassir
  Navab, and Slobodan Ilic.
\newblock 3d pictorial structures for multiple human pose estimation.
\newblock In {\em Proceedings of the IEEE Conference on Computer Vision and
  Pattern Recognition}, pages 1669--1676, 2014.

\bibitem{belagiannis20153d}
Vasileios Belagiannis, Sikandar Amin, Mykhaylo Andriluka, Bernt Schiele, Nassir
  Navab, and Slobodan Ilic.
\newblock 3d pictorial structures revisited: Multiple human pose estimation.
\newblock {\em IEEE transactions on pattern analysis and machine intelligence},
  38(10):1929--1942, 2016.

\bibitem{belagiannis2014multiple}
Vasileios Belagiannis, Xinchao Wang, Bernt Schiele, Pascal Fua, Slobodan Ilic,
  and Nassir Navab.
\newblock Multiple human pose estimation with temporally consistent 3d
  pictorial structures.
\newblock In {\em European Conference on Computer Vision Workshop}, pages
  742--754. Springer, 2014.

\bibitem{berclaz2011multiple}
Jerome Berclaz, Francois Fleuret, Engin Turetken, and Pascal Fua.
\newblock Multiple object tracking using k-shortest paths optimization.
\newblock {\em IEEE transactions on pattern analysis and machine intelligence},
  33(9):1806--1819, 2011.

\bibitem{Bewley2016_sort}
Alex Bewley, Zongyuan Ge, Lionel Ott, Fabio Ramos, and Ben Upcroft.
\newblock Simple online and realtime tracking.
\newblock In {\em 2016 IEEE International Conference on Image Processing
  (ICIP)}, pages 3464--3468, 2016.

\bibitem{bridgeman2019multi}
Lewis Bridgeman, Marco Volino, Jean-Yves Guillemaut, and Adrian Hilton.
\newblock Multi-person 3d pose estimation and tracking in sports.
\newblock In {\em Proceedings of the IEEE Conference on Computer Vision and
  Pattern Recognition Workshops}, 2019.

\bibitem{burenius20133d}
Magnus Burenius, Josephine Sullivan, and Stefan Carlsson.
\newblock 3d pictorial structures for multiple view articulated pose
  estimation.
\newblock In {\em Proceedings of the IEEE Conference on Computer Vision and
  Pattern Recognition}, pages 3618--3625, 2013.

\bibitem{cao2017realtime}
Zhe Cao, Tomas Simon, Shih-En Wei, and Yaser Sheikh.
\newblock Realtime multi-person 2d pose estimation using part affinity fields.
\newblock In {\em Proceedings of the IEEE Conference on Computer Vision and
  Pattern Recognition}, pages 7291--7299, 2017.

\bibitem{chen2019aggregate}
Long Chen, Haizhou Ai, Rui Chen, and Zijie Zhuang.
\newblock Aggregate tracklet appearance features for multi-object tracking.
\newblock {\em IEEE Signal Processing Letters}, 26(11):1613--1617, 2019.

\bibitem{chen2018cascaded}
Yilun Chen, Zhicheng Wang, Yuxiang Peng, Zhiqiang Zhang, Gang Yu, and Jian Sun.
\newblock Cascaded pyramid network for multi-person pose estimation.
\newblock In {\em Proceedings of the IEEE Conference on Computer Vision and
  Pattern Recognition}, pages 7103--7112, 2018.

\bibitem{chengocclusion}
Yu Cheng, Bo Yang, Bo Wang, Wending Yan, and Robby~T Tan.
\newblock Occlusion-aware networks for 3d human pose estimation in video.
\newblock In {\em ICCV}, 2019.

\bibitem{dong2019fast}
Junting Dong, Wen Jiang, Qixing Huang, Hujun Bao, and Xiaowei Zhou.
\newblock Fast and robust multi-person 3d pose estimation from multiple views.
\newblock In {\em Proceedings of the IEEE Conference on Computer Vision and
  Pattern Recognition}, pages 7792--7801, 2019.

\bibitem{elhayek2015efficient}
Ahmed Elhayek, Edilson de Aguiar, Arjun Jain, Jonathan Tompson, Leonid
  Pishchulin, Micha Andriluka, Chris Bregler, Bernt Schiele, and Christian
  Theobalt.
\newblock Efficient convnet-based marker-less motion capture in general scenes
  with a low number of cameras.
\newblock In {\em Proceedings of the IEEE Conference on Computer Vision and
  Pattern Recognition}, pages 3810--3818, 2015.

\bibitem{ershadi2018multiple}
Sara Ershadi-Nasab, Erfan Noury, Shohreh Kasaei, and Esmaeil Sanaei.
\newblock Multiple human 3d pose estimation from multiview images.
\newblock {\em Multimedia Tools and Applications}, 77(12):15573--15601, 2018.

\bibitem{hartley2003multiple}
Richard Hartley and Andrew Zisserman.
\newblock {\em Multiple view geometry in computer vision}.
\newblock Cambridge university press, 2003.

\bibitem{iskakov2019learnable}
Karim Iskakov, Egor Burkov, Victor Lempitsky, and Yury Malkov.
\newblock Learnable triangulation of human pose.
\newblock In {\em ICCV}, 2019.

\bibitem{joo2015panoptic}
Hanbyul Joo, Hao Liu, Lei Tan, Lin Gui, Bart Nabbe, Iain Matthews, Takeo
  Kanade, Shohei Nobuhara, and Yaser Sheikh.
\newblock Panoptic studio: A massively multiview system for social motion
  capture.
\newblock In {\em Proceedings of the IEEE International Conference on Computer
  Vision}, pages 3334--3342, 2015.

\bibitem{kuhn1955hungarian}
Harold~W Kuhn.
\newblock The hungarian method for the assignment problem.
\newblock {\em Naval research logistics quarterly}, 2(1-2):83--97, 1955.

\bibitem{lee2018propagating}
Kyoungoh Lee, Inwoong Lee, and Sanghoon Lee.
\newblock Propagating lstm: 3d pose estimation based on joint interdependency.
\newblock In {\em Proceedings of the European Conference on Computer Vision
  (ECCV)}, pages 119--135, 2018.

\bibitem{martinez2017simple}
Julieta Martinez, Rayat Hossain, Javier Romero, and James~J Little.
\newblock A simple yet effective baseline for 3d human pose estimation.
\newblock In {\em Proceedings of the IEEE International Conference on Computer
  Vision}, pages 2640--2649, 2017.

\bibitem{mehta2018single}
Dushyant Mehta, Oleksandr Sotnychenko, Franziska Mueller, Weipeng Xu, Srinath
  Sridhar, Gerard Pons-Moll, and Christian Theobalt.
\newblock Single-shot multi-person 3d pose estimation from monocular rgb.
\newblock In {\em 2018 International Conference on 3D Vision (3DV)}, pages
  120--130. IEEE, 2018.

\bibitem{mehta2017vnect}
Dushyant Mehta, Srinath Sridhar, Oleksandr Sotnychenko, Helge Rhodin, Mohammad
  Shafiei, Hans-Peter Seidel, Weipeng Xu, Dan Casas, and Christian Theobalt.
\newblock Vnect: Real-time 3d human pose estimation with a single rgb camera.
\newblock {\em ACM Transactions on Graphics (TOG)}, 36(4):1--14, 2017.

\bibitem{milan2016mot16}
Anton Milan, Laura Leal-Taix{\'e}, Ian Reid, Stefan Roth, and Konrad Schindler.
\newblock Mot16: A benchmark for multi-object tracking.
\newblock {\em arXiv preprint arXiv:1603.00831}, 2016.

\bibitem{moon2019camera}
Gyeongsik Moon, Ju~Yong Chang, and Kyoung~Mu Lee.
\newblock Camera distance-aware top-down approach for 3d multi-person pose
  estimation from a single rgb image.
\newblock In {\em ICCV}, 2019.

\bibitem{moreno20173d}
Francesc Moreno-Noguer.
\newblock 3d human pose estimation from a single image via distance matrix
  regression.
\newblock In {\em Proceedings of the IEEE Conference on Computer Vision and
  Pattern Recognition}, pages 2823--2832, 2017.

\bibitem{pavlakos2017harvesting}
Georgios Pavlakos, Xiaowei Zhou, Konstantinos~G Derpanis, and Kostas
  Daniilidis.
\newblock Harvesting multiple views for marker-less 3d human pose annotations.
\newblock In {\em Proceedings of the IEEE conference on computer vision and
  pattern recognition}, pages 6988--6997, 2017.

\bibitem{pavllo20193d}
Dario Pavllo, Christoph Feichtenhofer, David Grangier, and Michael Auli.
\newblock 3d human pose estimation in video with temporal convolutions and
  semi-supervised training.
\newblock In {\em Proceedings of the IEEE Conference on Computer Vision and
  Pattern Recognition}, pages 7753--7762, 2019.

\bibitem{qiu2019cross}
Haibo Qiu, Chunyu Wang, Jingdong Wang, Naiyan Wang, and Wenjun Zeng.
\newblock Cross view fusion for 3d human pose estimation.
\newblock In {\em ICCV}, 2019.

\bibitem{rayat2018exploiting}
Mir Rayat Imtiaz~Hossain and James~J Little.
\newblock Exploiting temporal information for 3d human pose estimation.
\newblock In {\em Proceedings of the European Conference on Computer Vision
  (ECCV)}, pages 68--84, 2018.

\bibitem{ristani2014tracking}
Ergys Ristani and Carlo Tomasi.
\newblock Tracking multiple people online and in real time.
\newblock In {\em Asian conference on computer vision}, pages 444--459.
  Springer, 2014.

\bibitem{tang2018joint}
Zheng Tang, Renshu Gu, and Jenq-Neng Hwang.
\newblock Joint multi-view people tracking and pose estimation for 3d scene
  reconstruction.
\newblock In {\em 2018 IEEE International Conference on Multimedia and Expo
  (ICME)}. IEEE, 2018.

\bibitem{taylor2010dynamical}
Graham~W Taylor, Leonid Sigal, David~J Fleet, and Geoffrey~E Hinton.
\newblock Dynamical binary latent variable models for 3d human pose tracking.
\newblock In {\em 2010 IEEE Computer Society Conference on Computer Vision and
  Pattern Recognition}, pages 631--638. IEEE, 2010.

\bibitem{tome2018rethinking}
Denis Tome, Matteo Toso, Lourdes Agapito, and Chris Russell.
\newblock Rethinking pose in 3d: Multi-stage refinement and recovery for
  markerless motion capture.
\newblock In {\em 2018 International Conference on 3D Vision (3DV)}, pages
  474--483. IEEE, 2018.

\bibitem{xiao2018simple}
Bin Xiao, Haiping Wu, and Yichen Wei.
\newblock Simple baselines for human pose estimation and tracking.
\newblock In {\em Proceedings of the European Conference on Computer Vision
  (ECCV)}, pages 466--481, 2018.

\bibitem{zhong2018camera}
Zhun Zhong, Liang Zheng, Zhedong Zheng, Shaozi Li, and Yi Yang.
\newblock Camera style adaptation for person re-identification.
\newblock In {\em Proceedings of the IEEE Conference on Computer Vision and
  Pattern Recognition}, pages 5157--5166, 2018.

\end{thebibliography}
}

\end{document}